\documentclass{article} 
\usepackage{iclr2023_conference,times}

\usepackage{amsmath,amsfonts,bm}

\def\eqref#1{equation~\ref{#1}}

\def\1{\bm{1}}

\DeclareMathAlphabet{\mathsfit}{\encodingdefault}{\sfdefault}{m}{sl}
\SetMathAlphabet{\mathsfit}{bold}{\encodingdefault}{\sfdefault}{bx}{n}

\usepackage{hyperref}
\usepackage{url}
\usepackage[pdftex]{graphicx}
\usepackage{microtype}
\usepackage{graphicx}
\usepackage{subcaption}
\usepackage{booktabs} 
\usepackage{dsfont}
\usepackage{amsfonts}
\usepackage{mathtools}
\usepackage{fancyhdr}
\usepackage{color}
\usepackage{algorithm}
\usepackage{algorithmic}
\usepackage{epsfig}
\usepackage{xspace}
\usepackage{multirow}
\usepackage{array}
\usepackage{comment}
\usepackage{amsthm}
\usepackage{makecell}
\usepackage{wrapfig}
\usepackage{bbm}

\usepackage{etoolbox}
\makeatletter
\AfterEndEnvironment{algorithm}{\let\@algcomment\relax}
\AtEndEnvironment{algorithm}{\kern2pt\hrule\relax\vskip3pt\@algcomment}
\let\@algcomment\relax
\newcommand\algcomment[1]{\def\@algcomment{\footnotesize#1}}
\newcommand{\ie}[1]{\textit{i.e.,}}

\renewcommand\fs@ruled{\def\@fs@cfont{\bfseries}\let\@fs@capt\floatc@ruled
	\def\@fs@pre{\hrule height.8pt depth0pt \kern2pt}%
	\def\@fs@post{}%
	\def\@fs@mid{\kern2pt\hrule\kern2pt}%
	\let\@fs@iftopcapt\iftrue}
\makeatother

\newcolumntype{I}{!{\vrule width 1pt}}
\newcommand{\stdv}[1]{\textcolor{darkgray}{\tiny$\pm$#1}}
\newcommand{\eg}[1]{\textit{e.g.,}}
\definecolor{myy}{RGB}{126,95,0}
\definecolor{mygray}{gray}{.9}
\definecolor{bblue}{RGB}{30,80,120}
\definecolor{mygray1}{gray}{.7}
\definecolor{ggray}{RGB}{127,127,127}

\def\1{\mathbbm{1}}
\makeatletter\renewcommand\paragraph{\@startsection{paragraph}{4}{\z@}
	{.5em \@plus1ex \@minus.2ex}{-.5em}{\normalfont\normalsize\bfseries}}\makeatother

\definecolor{mygreen}{HTML}{39b54a}  

\makeatletter
\newcommand{\thickhline}{%
	\noalign {\ifnum 0=`}\fi \hrule height 1pt
	\futurelet \reserved@a \@xhline
}
\makeatother

\newcommand{\xhdr}[1]{\vspace{1.7mm}\noindent{{\bf #1.}}}

\title{Prompt-driven efficient Open-set Semi-supervised Learning}

\author{Haoran Li$^{1}$, Chun-Mei Feng$^{2}$, Tao Zhou$^{4}$, Yong Xu$^{3}$, Xiaojun Chang$^{1}$ \\
$^{1}$AAII, Uniervisty of Technology Sydney, $^{2}$A*STAR, $^{3}$HIT (Shenzhen), $^{4}$NJUST \\
\texttt{\{haoran.li-3\}}@student.uts.edu.au}

\iclrfinalcopy 
\begin{document}

\maketitle

\begin{abstract}


Open-set semi-supervised learning (OSSL) has attracted growing interest, which investigates a more practical scenario where out-of-distribution (OOD) samples are only contained in unlabeled data. 
Existing OSSL methods like OpenMatch learn an OOD detector to identify outliers, which often update all modal parameters (\ie, full fine-tuning) to propagate class information from labeled data to unlabeled ones. 
Currently, prompt learning has been developed to bridge gaps between pre-training and fine-tuning, which shows higher computational efficiency in several downstream tasks.
In this paper, we propose a prompt-driven efficient OSSL framework, called \textit{OpenPrompt}, which can propagate class information from labeled to unlabeled data with only a small number of trainable parameters. We propose a prompt-driven joint space learning mechanism to detect OOD data by maximizing the distribution gap between ID and OOD samples in unlabeled data, thereby our method enables the outliers to be detected in a new way. The experimental results on three public datasets show that \textit{OpenPrompt} outperforms state-of-the-art methods with less than $1\%$ of trainable parameters. More importantly, \textit{OpenPrompt} achieves a $4\%$ improvement in terms of AUROC on outlier detection over a fully supervised model on CIFAR10.

\end{abstract}

\section{Introduction}
The goal of semi-supervised learning (SSL) is to improve a model's performance by leveraging unlabeled data~\cite{sohn2020fixmatch}. It can significantly improve recognition accuracy by propagating the class information from a small set of labeled data to a large set of unlabeled data without additional annotation cost~\cite{Li_2021_ICCV, Wang_2021_CVPR}. Existing SSL methods are built on the assumption that labeled and unlabeled data share the same distribution space. However, due to how it was collected, unlabeled data may contain new categories, such as outliers, that are never seen in the labeled data~\cite{saito2021openmatch}, resulting in lower SSL performance. To address this issue, Open Set SSL (OSSL) is proposed, the task of which is to classify in-distribution (ID) samples into the correct class while identifying out-of-distribution (OOD) samples as outliers~\cite{yu2020multi}.

Typical OSSL methods like MTC~\cite{yu2020multi} use a joint optimization framework to update the network parameters and the OOD score alternately. OpenMatch~\cite{saito2021openmatch} uses an OVA network that can learn a threshold to distinguish OOD samples from ID samples. Both of them follow the training strategy of current SSL methods, which \textbf{1)} propagate class information from a small set of labeled data to a large set of unlabeled data by fine-tuning all model parameters. Then use \textbf{2)} an additional structured OOD detector to identify outliers unseen in the labeled data that the unlabeled data may contain. However, such a mechanism leads to expensive computational costs.


As a new paradigm, prompting has shown outstanding effects in NLP~\cite{radford2021learning,shin2020autoprompt,jiang2020can}, which can make the model directly applicable to downstream tasks without introducing new parameters. Recently, prompting has been applied to computer vision tasks~\cite{jia2022vpt,bahng2022visual,sung2022lst}, which can greatly reduce the number of trainable parameters by modifying a small number of pixels to guide frozen vision models to solve new tasks. However, when the new task has OOD samples that have never appeared in the labeled data without supervision, the effectiveness of the prompt remains to be verified.

\begin{figure*}[t]
    \centering
    \includegraphics[width=0.88\textwidth]{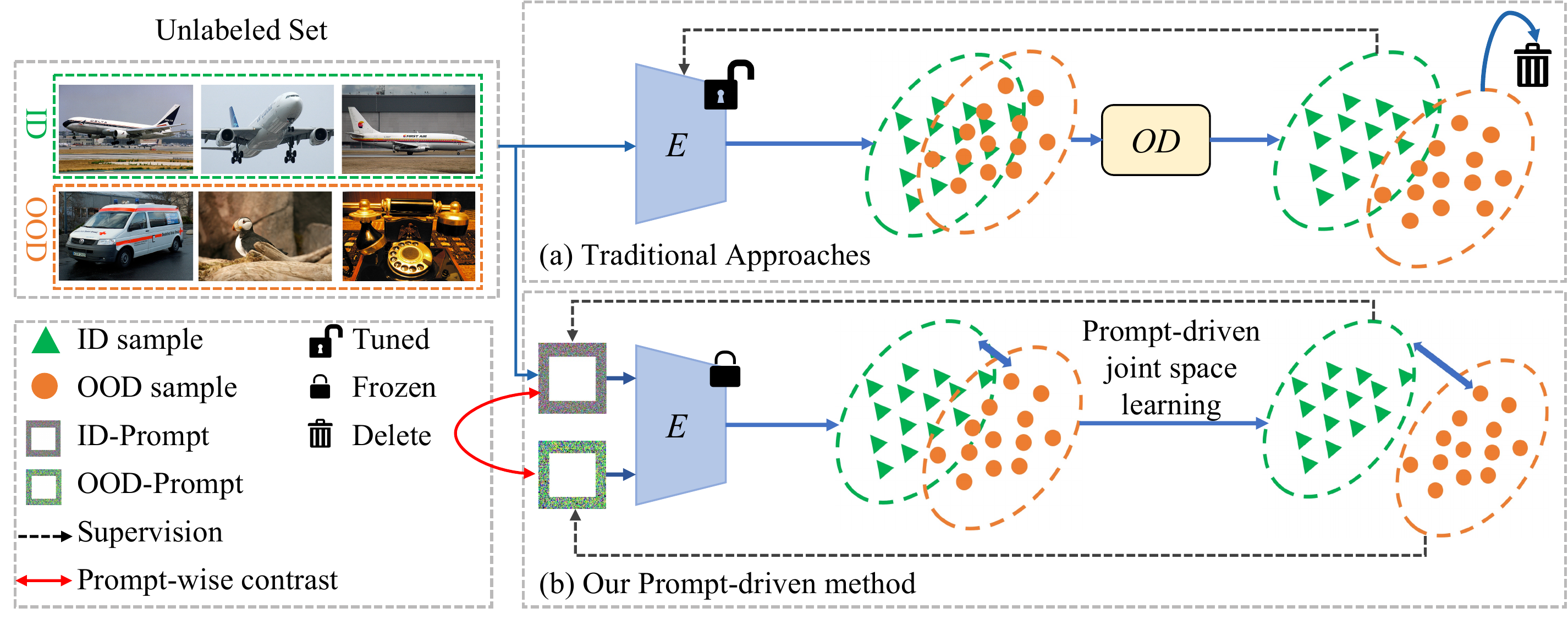} \vspace{-0.25cm}
    \caption{\textbf{Comparison of (a) traditional OSSL approaches and (b) our prompt-driven efficient OSSL,} where $E$ represents visual-encoder and $OD$ denotes outlier detector. (b) achieves class information propagation and OOD detection through only a small number of learnable parameters with a joint space learning mechanism, while (a) needs to fully fine-tune the whole model and an additional OOD detector. Different from existing works (a) which discard OOD samples after detection, (b) our method reuses them to train an OOD-specific prompt to push the OOD samples away from the ID ones.}
    \vspace{-0.5cm}
    \label{fig:motivation}
\end{figure*}

To this end, our study starts with the visual prompt-based OSSL task, where we hope to achieve \textbf{1)} class information propagation and \textbf{2)} OOD detection only by modifying a few pixels. First, we propose a prompt-driven efficient OSSL framework (\emph{\textit{OpenPrompt}}), which can effectively detect the outliers from unlabeled data with a small number of trainable parameters. In order to detect outliers without supervision, we project the representations of all samples into a prompt-driven joint space, which enlarges the distribution gap between ID and OOD samples (see Fig.~\ref{fig:motivation}). Instead of directly discarding the detected OOD samples (see Fig.~\ref{fig:motivation} (a)), we make full use of the detected OOD samples and feed them into the network to construct OOD-specific prompts. With prompts trained through ID data serving as positive samples, the OOD-specific prompts are regarded as negative samples to push OOD data away from ID samples (see Fig.~\ref{fig:motivation} (b)). Such a prompt-wise contrastive representation can further shape the joint space by exploring the structural information of labeled ID and OOD samples.

Overall, our contributions can be summarized as follows: 

\begin{itemize}

\item 
  {We propose a novel and efficient prompt-driven OSSL framework (termed \emph{\textit{OpenPrompt}}), which can only update a small number of learnable parameters to match the performance of full fine-tuning methods.}

\item 
   {We develop a prompt-driven joint space learning strategy to enlarge the distribution gap between ID and OOD data by visual prompts without supervision.}

\item 
   {We utilize a prompt-wise contrastive representation, \ie, ID-specific and OOD-specific prompt, to further shape the joint space instead of discarding OOD samples after detection.}
   
\item Experimental results show that \textit{OpenPrompt} performs better than other SOTA methods with no more than $1\%$ trainable parameters. Besides, \textit{OpenPrompt} achieves the best performance in detecting outliers compared with other methods.
\end{itemize}


\section{Related Work}\label{Sec.2}
\xhdr{Semi-supervised learning (SSL)} The SSL methods assume the labeled and unlabeled data are from the same classes and propagate class information from labeled to unlabeled data to improve a model’s performance~\cite{rasmus2015semi, berthelot2019mixmatch, verma2022interpolation, zheng2022simmatch, french2019semi}. 
Existing pseudo-labeling~\cite{lee2013pseudo} and consistency regularization~\cite{laine2017temporal} based SSL approaches show great performance on many benchmark datasets. They usually set a selection threshold to make sure all pseudo-labels used are reliable. Different from previous methods, U$^{2}$PL~\cite{wang2022semi} treats unreliable pseudo-labels as negative samples for contrastive learning. Several methods~\cite{devries2017improved, yang2022st++, yun2019cutmix, yuan2021simple} utilize different data augmentations with self-training to maintain the consistency of the model. All these existing works often propagate class information from labeled to unlabeled data using a full fine-tuning adaption, however, it will require large computing resources by adapting the whole model to the unlabeled data. 

\xhdr{Open-set Semi-supervised learning (OSSL)} OSSL relaxes the SSL assumption by presuming that unlabeled data contains instances from novel classes, which will lower the performance of SSL approaches. Existing OSSL approaches use outlier detectors to recognize and filter those instances to improve the model's robustness. MTC introduces a joint optimization framework, which updates the network and the discriminating score of unlabeled data alternately~\cite{yu2020multi}. UASD generates soft targets as regularisers to empower the robustness of the proposed SSL network~\cite{chen2020semi}. OpenMatch applies a soft consistency loss to the outlier detector by outputting the confidence score of a sample to detect outliers~\cite{saito2021openmatch}. By contrast, our \textit{OpenPrompt} uses visual prompts to distinguish the representation gaps between ID and OOD samples and applies prompt-driven joint space learning to enlarge these gaps for outlier detection. Experimental results demonstrate that our approach achieves great performance through this prompt-driven joint space learning mechanism.

\xhdr{Prompting} Prompting is a \textit{modus operandi} method, which aims to adapt pre-trained models to downstream tasks by modifying the data space~\cite{radford2021learning}. With manually chosen prompt candidates, GPT-3 performs well on downstream transfer learning tasks~\cite{brown2020language}. Recent studies in computer vision treat prompts as task-specific vectors and by tuning prompts instead of the models' parameters, the pre-trained models could also be adapted to downstream tasks with the same performance. VPT injects learnable prompt tokens to the vision transformer and keeps the backbone frozen during the downstream fine-tuning stage, which achieves great performance with only a small amount of learnable parameters~\cite{jia2022vpt}. Bahng \textit{et al.} applies learnable paddings to the data space as a visual prompt which uses fewer parameters to realize large-scale model adaption~\cite{bahng2022visual}. Inspired by those recent studies on vision prompts, we proposed an effective prompt-based OSSL framework. Different from existing approaches, we apply a prompt for a robust adaption by rejecting noisy samples during the downstream fine-tuning stage. 
\vspace{-0.3cm}

\section{Methods}\label{Sec.3}



\textbf{Problem setting}. Our task is to train an effective classification network via OSSL. In OSSL, we divide the used dataset into two parts: the labeled data $\mathcal{X}_l = \{(x_i,y_i)\}_{i=1}^n$ and unlabeled data $\mathcal{X}_u = \{(x_i\}_{i=1}^m$, and the class spaces of labeled and unlabeled data are denoted as $\mathcal{O}_l$ and $\mathcal{O}_u$, respectively. In this study, we assume that $\mathcal{O}_l \subset \mathcal{O}_u$ and $\mathcal{O}_l \ne \mathcal{O}_u$. Besides, the unlabeled data that belong to the class space $\mathcal{O}_l$ are called in-distribution (ID) samples, while the unlabeled data only belonging to the label space $\mathcal{O}_u$ are called out-of-distribution (OOD) samples. Therefore, the goal of OSSL is to train a robust model to classify ID samples into the correct classes while learning to effectively detect OOD samples.


\subsection{Overview of \textit{OpenPrompt}}\label{Sec.3.2}

The main challenge of this task is to find an efficient way to train a robust network when the labels and categories of the used dataset are both imbalanced. Inspired by recent studies~\cite{jia2022vpt,bahng2022visual} on visual prompts, we propose a prompt-driven efficient OSSL framework (termed \textit{OpenPrompt}), to improve the efficiency of OSSL task.
In detail, the proposed framework includes two stages, \ie, conducting pre-training on labeled data and fine-tuning on unlabeled data. 
In this way, a visual prompt is inserted into the data space~\cite{bahng2022visual}, which can propagate class information from labeled to unlabeled data with only a small amount of trainable parameters while keeping the model frozen. Moreover, instead of introducing additional structures as the outlier detector, we propose a prompt-driven joint space learning mechanism to detect OOD samples.


To make full use of the detected OOD samples, we propose a prompt-wise contrastive representation strategy, which can further enlarge the distribution gap between ID and OOD samples. Instead of discarding the OOD samples as other OSSL approaches~\cite{saito2021openmatch,yu2020multi}, we feed them into the network and train an OOD-specific visual prompt as negative samples for a prompt-wise contrastive representation.

\begin{figure*}[t]
    \centering
    \includegraphics[width=0.9\textwidth]{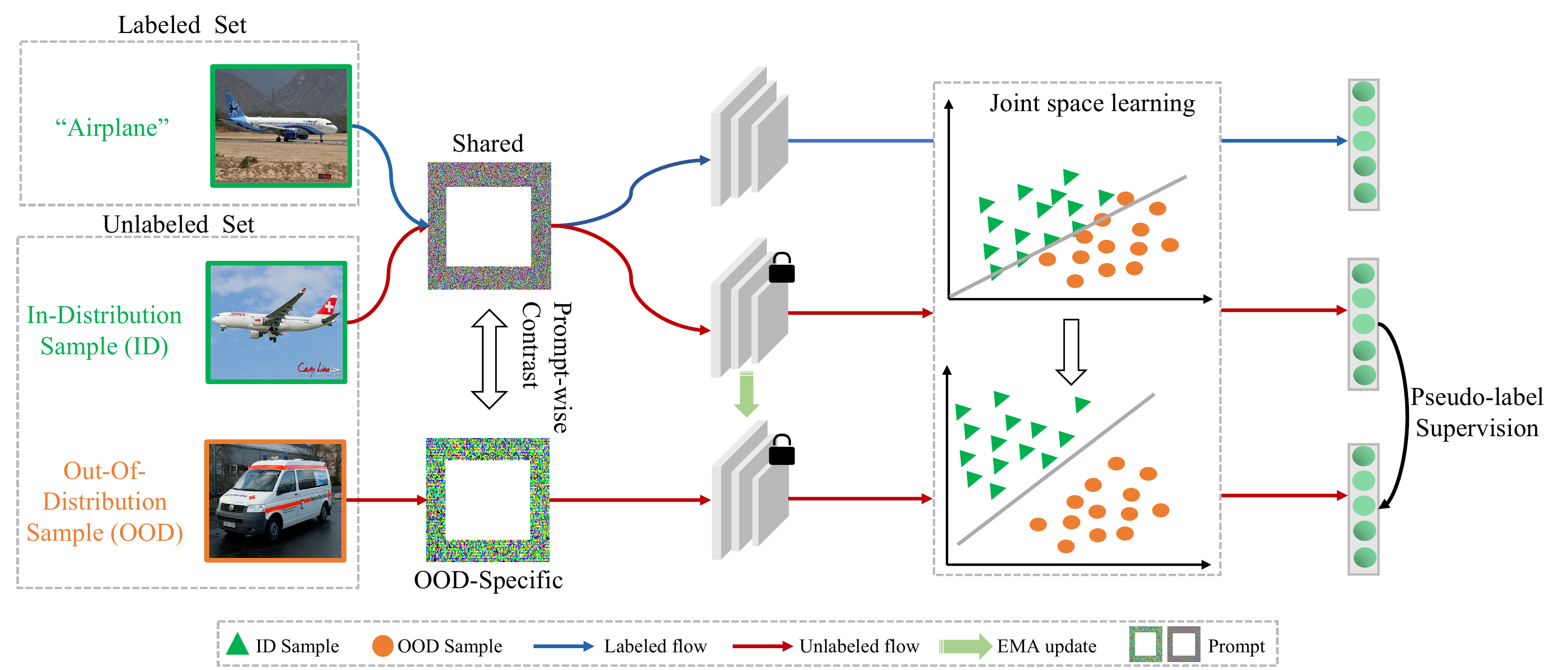} 
    \caption{Overview of our proposed framework. In unlabeled flow, two different augmented inputs are fed into the network to obtain classification results. The joint space learning mechanism is used to enlarge the distribution difference distinguished by the shared ID-prompt for OOD detection. The pseudo-label supervision only uses confident pseudo-labels, \ie, a threshold is set for selection.
    }
    \vspace{-0.5cm}
    \label{fig:framework}
\end{figure*}

As shown in~Fig.~\ref{fig:framework}, our framework utilizes a Teacher-Student structure, including a teacher model and a student model. Each model has four components: (1) a visual-encoder $E(\cdot)$, (2) a learnable visual prompt $v_{\phi}$ parameterized by $\phi$ in the form of pixels, (3) a prompt-driven joint space $S(\cdot)$, and (4) a closed-set classifier $C(\cdot)$. 

In the pre-training stage, $\mathcal{X}_l = \{(x_i,y_i)\}_{i=1}^n$ are feed into the $E(\cdot)$ with the prompt $v_{\phi}$ to obtain the feature representations $F_l = \{f_i \in \mathbb{R}^D\}_{i=1}^n$, which are then feed into the classifier $C(\cdot)$ to obtain the prediction results $\hat{Y} = \{\hat{y}_i\}_{i=1}^n$. The labeled objective function $\mathcal{L}_l$ is used to infer the loss with the ground-truth ${Y} = \{{y}_i\}_{i=1}^n$. 
Noted that, all parameters are set to be learned in this stage for training a reliable classification model and a visual prompt. 
Since there are no OOD samples contained during this stage, it can not find an appropriate binary classifier in the joint space to distinguish ID and OOD data. Therefore, we set multiple binary-classifier candidates in this stage, which are used to be selected in the fine-tuning stage. See details of the candidate selection in \S\ref{Sec.3.3}. 

In the fine-tuning stage, we feed $\mathcal{X}_u = \{x_i\}_{i=1}^m$ into both the teacher and student networks with different augmentations, 
to obtain the feature representations $F_u = \{f_i \in \mathbb{R}^D\}_{i=1}^m$ for unlabeled data. The joint space takes $F_u$ as inputs and determines whether they belong to ID or OOD samples. Features of ID samples will be sent to the classifier and the final class prediction is calculated by $\hat{Y} = \{\hat{y}_i\}_{i=1}^m= C (F_u)$. Features of OOD samples will not be used for classification and all OOD samples will be resent to the network with an initialized OOD-specific visual prompt $\overline{v_{\phi}}$ 
to train an OOD-specific prompt as negative samples for contrastive learning. The supervision signals used for the unlabeled objective function $\mathcal{L}_u$ are pseudo labels received from the student network, and the classification results are from the teacher network. Noted that in this stage, only prompt $v_{\phi}$ and OOD-specific prompt $\overline{v_{\phi}}$ are updated through the objective function while the remaining network remains frozen. Following~\cite{tarvainen2017mean}, the student network is updated via optimizing $\mathcal{L}_l$ and $\mathcal{L}_u$, while the teacher network is updated via exponential moving average (EMA).

The labeled objective function $\mathcal{L}_l$ only contains a supervised loss, while the unlabeled objective function contains three components, \ie, \textit{(i)} supervised objective, \textit{(ii)} consistency objective, and \textit{(iii)} contrastive learning objective, which can be formulated by
\begin{align}
\mathcal{L}_u = \mathcal{L}_{\text{S}}+\mathcal{L}_{\text{C}}+\mathcal{L}_{\text{CL}},
\end{align}
where $\mathcal{L}_{\text{S}}$ denotes supervised loss, $\mathcal{L}_{\text{C}}$ denotes consistent loss, and $\mathcal{L}_{\text{CL}}$ denotes contrastive learning loss which is calculated through the prompt $v_{\phi}$ and OOD-specific prompt $\overline{v_{\phi}}$.

\subsection{Prompt-driven Joint Space Learning}\label{Sec.3.3}
\subsubsection{Prompt design}
We introduce a parameterized visual prompt $v_{\phi}$ in the form of pixels to the input image $x$ and form a prompted input $x+v_{\phi}$. Following~\cite{bahng2022visual}, $v_{\phi}$ is designed using the padding template to achieve the best performance. Therefore, given the prompt size $p$, the actual number of parameters is $2Cp \times (H + W-2p)$, where $C$, $H$, and $W$ are the image channels, height, and width respectively.
\subsubsection{Joint space learning}

\xhdr{Pre-training stage} Different from~\cite{jia2022vpt,bahng2022visual}, we learn a visual prompt in the pre-training stage, which is used to propagate the class information of labeled data to unlabeled ones in the fine-tuning stage. Specifically, the prompted input $x+v_{\phi}$ is fed into the visual-encoder $E(\cdot)$ to obtain the output feature representations. Then, the feature representations $F_l = \{f_i \in \mathbb{R}^D\}_{i=1}^n$ are fed into the joint space $S(\cdot)$ to enlarge the distribution gap for OOD detection. We assume that all ID samples form a cluster $\mathcal{K}$ in the joint space, which can be defined by
\begin{equation}
\mathcal{K} = \{f_i, k_{ic}\}_{i=1}^n, 
\end{equation}
where $k_{ic}$ denotes the cluster center of the ID samples. It is critical to initialize a binary classifier for OOD detection. We build a circle through the ID cluster $\mathcal{K}$ and use its tangent lines as the initial binary classifiers. Given the cluster center $k_{ic}$, we define the radius $r$ of the circle as follows:
\begin{equation}
r = \mathop{\max}_{i \in n} d(f_i, k_{ic}), 
\end{equation}
where $d(\cdot , \cdot)$ denotes an Euclidean distance. To ensure all ID samples are correctly detected, we choose the furthest sample and use the distance between it and the cluster center as the radius. As mentioned in \S\ref{Sec.3.2}, multiple binary classifier candidates are set for selection in the fine-tuning stage. To achieve this, we choose the top $N$ furthest samples from the cluster center and build tangent lines on each of them. Since the furthest sample is used to form the circle, we have $N-1$ samples which are not on the boundary of the circle. As the tangent line of a circle must pass through a point on the boundary of the circle, we push these $N-1$ points outward some distance to ensure that they lie on the boundary of the circle. Finally, the initial classifier candidates can be defined by
\begin{equation}
D_j(f_i)=\left\{
\begin{array}{ll}
F(f_i), & j = 1 \vspace{1ex}\\
F(f_i) + (r - d(f_i, k_{ic})), & 1 < j \leq N
\end{array}
\right.
\end{equation}
where $F(\cdot)$ denotes the tangent function~\cite{thurston1964definition}. It is worth noting that we can obtain promising performance when $N=5$ as discussed in \S\ref{Sec.4.2}.

\xhdr{Fine-tuning stage} The parameterized visual prompt is also added to the unlabeled input image in this stage. It is worth noting that unlike others~\cite{jia2022vpt} which learn a prompt from an initialized one in this stage, our method here inherits the visual prompt $v_{\phi}$ learned from the pre-training stage. 
Given feature representations $F_u = \{f_i \in \mathbb{R}^D\}_{i=1}^m$ obtained through the prompted unlabeled images, we then feed $F_u$ into the joint space $S(\cdot)$ for OOD detection. Since the labeled and unlabeled ID samples share similar classes, the outputs of $D_j(F_l)$ should be close to those of $D_j(F_u)$ from ID data. Therefore, we can detect OOD data by:
\begin{equation}
\left\{
\begin{array}{ll}
x_i \text{ is ID sample } x_{id}, & \text {if } \mid D_j(f_i) - \overline{D_j(F_l)} \mid\leq 0.1 \vspace{1ex}\\
x_i \text{ is OOD sample } x_{ood}, & \text {otherwise}
\end{array}
\right.
\end{equation}

\begin{wrapfigure}{r}{0.5\textwidth}
\begin{center}
\vspace{-0.2cm}
\includegraphics[width=0.4\textwidth]{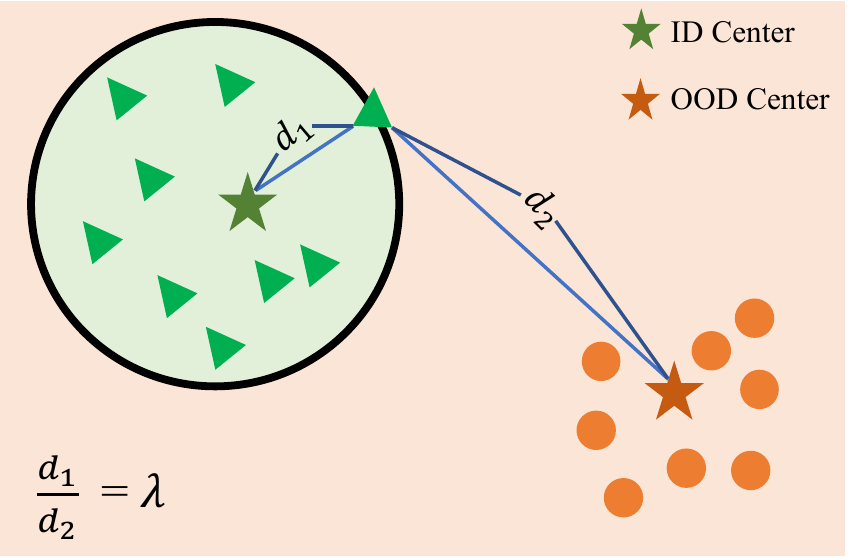}
  \end{center}
  \vspace{-0.2cm}
  \caption{Apollonius circle~\cite{kim2009circle} built on two centers. Definition of Apollonius circle is given in Appendix~\ref{ap:1}.}
  \label{fig:AC}
  \vspace{-0.4cm}
\end{wrapfigure}
where $\overline{D_j(F_l)}$ denotes the average of the outputs of $D_j(F_l)$.

Then we can get the rate $r_j$ of OOD samples in unlabeled data using the $j$-th binary classifier candidate:
\begin{equation}
r_j = \frac{n_{j}}{m}, 
\end{equation}
where $n_{j}$ denotes the number of detected OOD samples. The candidate with the largest rate will be selected as the initial binary classifier $D(\cdot)$ for both the student and teacher networks. 

Then we resend $F_u = \{f_i \in \mathbb{R}^D\}_{i=1}^m$ back to the joint space $S(\cdot)$ in both two networks to conduct OOD detection and update the binary classifier. Due to the binary classifier being built on ID samples from labeled data, which will cause it biased toward them, we need to modify the binary classifier. The linear classifier can be justified through a calibrated stacking method~\cite{chao2016empirical}, which directly shifts the decision boundary through a fixed value. However, this will degrade the performance in detecting ID samples. Inspired by~\cite{baek2021exploiting}, instead of shifting, we modulate the decision boundary with the Apollonius circle. Specifically, we first use the initial $D(\cdot)$ to detect OOD samples from unlabeled data and calculate the initial cluster center $k_{oc}$ of these samples. Then, given a feature $f_i$ from $F_u$, we can compute two distances from $k_{ic}$ and $k_{oc}$ to $f_i$ by
\begin{equation}
d_1 = d(f_i, k_{ic}) \mbox{ and } d_2  = d(f_i, k_{oc}),
\end{equation}
where $d_1 \leq d_2$. Therefore, the Apollonius circle used as the binary-classifier can be formulated with an adjustable parameter $\lambda$ as follows:
\begin{equation}
\mathcal{A}(\lambda) = \{f_i \mid d1:d2 = \lambda\},
\end{equation}
where $\mathcal{A}(\lambda)$ denotes the boundary of the Apollonius circle (see Fig~\ref{fig:AC}). Therefore, the decision rule of the binary classifier is modulated as follows:
\begin{equation}
D_{ac}(f_i)=c_\mathrm{id}\boldsymbol{\mathbbm{1}}\left[\frac{d_1}{d_2} \leq \lambda\right] + 
	c_\mathrm{ood}\boldsymbol{\mathbbm{1}}\left[\frac{d_1}{d_2} > \lambda\right],
\end{equation}
where $c_\mathrm{id}$ or $c_\mathrm{ood}$ denotes an identification that is ID or OOD sample, and $\mathbbm{1}$ denotes an indicator function that returns 1 if the argument is true, and 0 otherwise. Noted that two cluster centers $k_{ic}$ and $k_{oc}$ are updated through the new detected ID and OOD samples after each detection process. After that, we then send all ID samples to the classifier $C(\cdot)$ for prediction and use the detected OOD samples for prompt-wise contrastive representation learning (as mentioned in \S\ref{Sec.3.3}). 

\xhdr{Prompt-wise contrastive representation} Existing approaches often filter out the OOD samples once detect them. We notice that those unused data could be used to train an OOD-specific prompt as negative samples to further improve the model performance. Therefore, at the end of each training epoch, we collect all the detected OOD samples and resend them to the network with a new initial visual prompt. We call this prompt an OOD-specific prompt $\overline{v_{\phi}}$ and treat them as negative samples to apply contrastive learning.

\subsection{Objective function}\label{Sec.3.5}
As discussed in \S\ref{Sec.3.2}, our overall objective function includes supervised loss $\mathcal{L}_{\text{S}}$, consistent loss $\mathcal{L}_{\text{C}}$, and contrastive learning loss $\mathcal{L}_{\text{CL}}$.

\xhdr{Supervised loss} Following existing classification works, we use cross-entropy (CE) loss $\mathcal{L}_{CE}$ as the supervised loss. Noted that for the pre-training stage, we directly computer the $\mathcal{L}_{CE}$ through ground-truth $\mathcal{L}_{\text{CE}} (Y, \hat{Y})$. However, for the fine-tuning stage, there are no ground-truth labels used as supervision signals. Therefore, we set the prediction results from student network as pseudo-labels $\check{Y} = \{\check{y}_i\}_{i=1}^n$ for supervision signals and use a threshold~\cite{sohn2020fixmatch} $\eta = 0.7$ to make sure all used pseudo-labels are reliable.

\xhdr{Consistent loss} For consistent loss, different from existing works which calculate the per pixel-level consistency of each feature map, we calculate the consistency through prompting and joint space learning. As mentioned in \S\ref{Sec.3.2}, we conduct different augmentation strategies on the input samples. Since two different augmented inputs use the same prompt, they should share similar feature representations. Therefore, we use the Euclidean distance in the joint space as the consistency loss:
\begin{equation}
\mathcal{L}_{\text{C}} = {d}(k_{ic}, f_{s}) + {d}(k_{oc}, f_{s}) - {d}(k_{ic}, f_{t}) - {d}(k_{oc}, f_{t}),
\end{equation}
where $f_{s}$ and $f_{t}$ denote features obtained from student and teacher networks, respectively. 

\xhdr{Contrastive learning loss} We resend OOD samples with a random initialized visual prompt $\overline{v_{\phi}}$ to both the student and teacher networks, and then use $\mathcal{L}_{\text{C}}$ for optimization to get the OOD-specific prompt. Since this prompt is only trained using OOD samples, we treat this one as a negative sample. Our goal is to enlarge the distribution gap between $v_{\phi}$ and $\overline{v_{\phi}}$. Therefore, the contrastive learning loss can be defined by
\begin{equation}
\mathcal{L}_{\text{CL}} = 1 - \mathcal{L}_{\text{COS}}(v_{\phi}, \overline{v_{\phi}}),
\end{equation}
where $\mathcal{L}_{\text{COS}}$ denotes the cosine similarity loss.
\section{Experiments}\label{Sec.4}

\begin{table}[t]
\renewcommand{\arraystretch}{1.0}
	\caption{Average (with standard deviation) AUROC results. Paras. denotes the total learnable parameters of the model based on Wide
ResNet-28-2 for CIFAR 10 and CIFAR100, and based on ResNet-18 for ImageNet-30. The number of classes for ID and OOD and the number of labeled samples for each class are shown in each column.}
\vspace{-2mm}
	\label{tab:benchmarks}
        \setlength{\tabcolsep}{3pt}
	\fontsize{7.6}{9}\selectfont
	\centering
	\begin{tabular}{l c c c ccc c cc c cc c c}
\toprule

Dataset &\multirow{3}*{\rotatebox{270}{Paras.(M)}}& \multicolumn{3}{c}{CIFAR10} && \multicolumn{2}{c}{CIFAR100} &&\multicolumn{2}{c}{CIFAR100}&& \multicolumn{1}{c}{ImageNet-30} \\
\cmidrule{1-1} \cmidrule{3-5} \cmidrule{7-8}  \cmidrule{10-11} \cmidrule{13-13} 
No. of ID / OOD  && \multicolumn{3}{c}{6 / 4} && \multicolumn{2}{c}{55 / 45} &&
\multicolumn{2}{c}{80 / 20} && 20 / 10\\
\cmidrule{1-1} \cmidrule{3-5} \cmidrule{7-8}  \cmidrule{10-11} \cmidrule{13-13}
No. of Labeled && \multirow{2}*{50} & \multirow{2}*{100} & \multirow{2}*{400}  &&\multirow{2}*{50} & \multirow{2}*{100} && \multirow{2}*{50} & \multirow{2}*{100} && \multirow{2}*{10 \%}\\samples\\
\midrule
\noalign{\smallskip}
Labeled Only&- / - &63.9\stdv{0.5}&64.7\stdv{0.5}&76.8\stdv{0.4}&&76.6\stdv{0.9}&79.9\stdv{0.9}&&70.3\stdv{0.5}&73.9\stdv{0.9}&&80.3\stdv{1.0}\\
FixMatch&23.83/33.22&56.1\stdv{0.6}&60.4\stdv{0.4}&71.8\stdv{0.4}&&72.0\stdv{1.3}&75.8\stdv{1.2}&&64.3\stdv{1.0}&66.1\stdv{0.5}&&88.6\stdv{0.5}\\
MTC&23.36/32.66&96.6\stdv{0.6}& 98.2\stdv{0.3}&98.9\stdv{0.1}&&	81.2\stdv{3.4}&80.7\stdv{4.6}&&79.4\stdv{2.5}&73.2\stdv{3.5}&&93.8\stdv{0.8}\\
OpenMatch &23.44/32.73&{99.3}\stdv{0.3}& 99.7\stdv{0.2}&	\bf{99.3}\stdv{0.2}&&87.0\stdv{1.1}&86.5\stdv{2.1}&&86.2\stdv{0.6}&86.8\stdv{1.4}&&96.4\stdv{0.7}\\
\midrule
$~\textbf{\texttt{Ours}}$ \textit{w/o} CL &\bf{0.08/0.08}&98.7\stdv{0.3}& 99.5\stdv{0.7}&	98.8\stdv{0.1}&&86.7\stdv{1.2}&86.4\stdv{0.2}&&86.0\stdv{0.5}&86.4\stdv{0.3}&&95.8\stdv{1.1}\\
$~\textbf{\texttt{Ours}}$  &\bf{0.16/0.16}&\bf{99.4}\stdv{0.1}& \bf{99.7}\stdv{0.5}&	99.2\stdv{0.1}&&\bf{87.2}\stdv{1.2}&\bf{87.0}\stdv{0.6}&&\bf{86.5}\stdv{1.5}&\bf{87.3}\stdv{0.4}&&\bf{97.1}\stdv{0.2}\\
\bottomrule
\vspace{-0.75cm}
\end{tabular}
\end{table}

\xhdr{Implementation Details} Following~\cite{saito2021openmatch}, we implemented our network based on Wide ResNet-28-2~\cite{zagoruyko2016wide} (for CIFAR10 and CIFAR100) and ResNet-18 (for ImageNet-30). The models are trained on one NVIDIA Titan X with a 12-GB GPU. We use the standard SGD with an initial learning rate of 0.3, and momentum set as 0.9. The hyperparameters prompt size $p$, candidate number $N$, and $\lambda$ are empirically set to 40, 5, and 0.5, separately (see discussions in \S\ref{Sec.4.2}). The average result of three runs and its standard deviation are recorded.

\xhdr{Datasets} \label{Sec.DATA}
The proposed \textit{OpenPrompt} is evaluated on three OSSL benchmark image classification datasets, including CIFAR-10, CIFAR-100~\cite{krizhevsky2009learning} and ImageNet-30~\cite{deng2009imagenet}. In the OSSL task, the test set is assumed to contain both known (ID) and unknown (OOD) classes~\cite{saito2021openmatch,yu2020multi}. Specifically, for CIFAR10, we use the animal classes (six classes) as ID data and the other four classes as OOD data. For CIFAR100, we have two experimental settings: 80 classes as ID data (20 classes as OOD data) and 55 classes as ID data (45 classes as OOD data)~\cite{saito2021openmatch}. For ImageNet-30, we pick the first 20 classes (in alphabetical order) as ID data and use the remaining 10 classes as OOD data~\cite{saito2021openmatch}.

\xhdr{Baselines} 
We use MTC~\cite{yu2020multi} and OpenMatch~\cite{saito2021openmatch} with the source codes as the OSSL baselines. Following~\cite{saito2021openmatch}, we train two models, one using only labeled samples (Labeled Only) and the other one employing the FixMatch~\cite{sohn2020fixmatch} method. Then, we add the outlier detector from OpenMatch to these two models for OOD detection. The hyper-parameters of all the methods are tuned by maximizing the AUROC on the validation set.

\subsection{Results}\label{Sec.4.1}
\xhdr{CIFAR10 and CIFAR100} Following MTC~\cite{yu2020multi} and OpenMatch~\cite{saito2021openmatch}, we use AUROC as the evaluation metric. The results are shown in Table~\ref{tab:benchmarks}, where the number of classes for ID and OOD and the number of labeled samples for each class are shown in each column. Since we employ the ResNet-28-2~\cite{he2016deep} network architecture for CIFAR10 and CIFAR100, and the ResNet-18 for ImageNet-30, the number of learnable parameters for each of the two basic network architectures are respectively recorded. As can be seen from this table, \textit{OpenPrompt} achieves the best performance in most cases on both CIFAR10 and CIFAR100. For example, \textit{OpenPrompt} increased the AUROC values from 86.4 to 87.3 in CIFAR100 at 100 labels. More importantly, the learnable parameters of our proposed \textit{OpenPrompt} are far less than the existing methods, \eg, $23.44 M$ $\rightarrow$ $\textbf{0.16 M}$ compared with OpenMatch and $23.8 M$ $\rightarrow$ $\textbf{0.16 M}$ compared with FixMatch~\cite{sohn2020fixmatch}, which are no more than $1\%$. This is because our method adapts a model on labeled data to unlabeled by modifying the data space while others modify the model space~\cite{bahng2022visual}.

\xhdr{ImageNet-30} Here, we also evaluate the proposed \textit{OpenPrompt} on the more challenging dataset ImageNet-30~\cite{saito2021openmatch}. As described in \S\ref{Sec.DATA}, we alphabetically select the first 20 classes as ID classes and the remaining 10 classes as OOD classes. As shown in the rightmost column of Table~\ref{tab:benchmarks}, \textit{OpenPrompt} also has the highest AUROC accuracy on ImageNet-30.
In particular, we see our proposed \textit{OpenPrompt} leading to substantial performance gains (\ie, $96.4$ $\rightarrow$ $\textbf{97.1}$) with 10 \% labels in ImageNet-30 compared to OpenMatch. All methods use the ResNet-18 network architecture on the ImageNet-30 dataset, and the learnable parameters are $33.2 M$, $32.66 M$, and $32.73 M$, respectively. However, our method freezes the entire network except for visual cues, and its learnable parameters are much smaller than various SOTA baselines, which are just 0.16 M. This indicates that \textit{OpenPrompt} achieves promising performance on complex and challenging datasets.

\begin{table}[t]
\renewcommand{\arraystretch}{0.90}
\caption{AUROC evaluation of OOD detection. The higher values, the better performance. Supervised models indicate our proposed OpenMatch but are trained with fully labeled data.}
\vspace{-0.3cm}
\setlength{\tabcolsep}{4pt}
\begin{subtable}{1\textwidth}
\caption{Training on CIFAR10, where there is 100 labeled data per class and unlabeled data. }
\centering
\resizebox{0.83\textwidth}{!}{
\begin{tabular}{lcccccc}
\toprule
Method  &CIFAR10& SVHN & LSUN & ImageNet & CIFAR100& MEAN \\
\midrule
Labeled Only&64.7\stdv{1.0}&83.6\stdv{1.0}&78.9\stdv{0.9}&80.5\stdv{0.8}&80.4\stdv{0.5}&80.8\stdv{0.8}\\
FixMatch~\cite{sohn2020fixmatch}&60.4\stdv{0.4}&79.9\stdv{1.0}&67.7\stdv{2.0}&76.9\stdv{1.1}&71.3\stdv{1.1}&73.9\stdv{1.3}\\
MTC~\cite{yu2020multi} &98.2\stdv{0.3}& 87.6\stdv{0.5}& 82.8\stdv{0.6}&96.5\stdv{0.1}&90.0\stdv{0.3}&89.2\stdv{0.4}\\
OpenMatch~\cite{saito2021openmatch} & {99.7}\stdv{0.1} & {93.0}\stdv{0.4}&{92.7}\stdv{0.3}&\bf{98.7}\stdv{0.1}&{95.8}\stdv{0.4}&{95.0}\stdv{0.3}\\
$~\textbf{\texttt{Ours}}$ & \bf{99.7}\stdv{0.2} & \bf{94.1}\stdv{1.1}&\bf{93.6}\stdv{0.7}&97.4\stdv{0.3}&\bf{96.2}\stdv{0.5}&\bf{95.4}\stdv{1.3}\\
\midrule
\midrule
Supervised & 89.4\stdv{1.0}& 95.6\stdv{0.5} &89.5\stdv{0.7} &90.8\stdv{0.4}&90.4\stdv{1.0}&91.6\stdv{0.6}\\
\bottomrule
\end{tabular}}
\label{tab:CIFAR10}
\end{subtable}

\begin{subtable}{\textwidth}
\vspace{1.5mm}
\caption{Training on ImageNet-30, where there is 10 \% of labeled data and unlabeled data. }
\vspace{-0.8mm}
\resizebox{\textwidth}{!}{
\begin{tabular}{lcccccccc}
\toprule
Method & ImageNet-30 &LSUN&DTD&CUB&Flowers&Caltech&Dogs&MEAN\\
\midrule
Labeled Only&80.3\stdv{0.5}&85.9\stdv{1.4} &75.4\stdv{1.0}&77.9\stdv{0.8}&69.0\stdv{1.5}&78.7\stdv{0.8}&84.8\stdv{1.0}&78.6\stdv{1.1}\\
FixMatch~\cite{sohn2020fixmatch}&88.6\stdv{0.5}&85.7\stdv{0.1}&83.1\stdv{2.5}&81.0\stdv{4.8}&{81.9}\stdv{1.1}&83.1\stdv{3.4}&86.4\stdv{3.2}&83.0\stdv{1.9}\\
MTC~\cite{yu2020multi} &93.8\stdv{0.8}&78.0\stdv{1.0}&59.5\stdv{1.5}&72.2\stdv{0.9}&76.4	\stdv{2.1}&80.9\stdv{0.9}&78.0\stdv{0.8}&74.2\stdv{1.2}\\
OpenMatch~\cite{saito2021openmatch}& {96.3}\stdv{0.7} & {89.9}\stdv{1.9}& {84.4}\stdv{0.5}& {87.7}\stdv{1.0}& 80.8\stdv{1.9}& {87.7}\stdv{0.9}& \bf{92.1}\stdv{0.4}& {87.1}\stdv{1.1}\\
$~\textbf{\texttt{Ours}}$& \bf{97.1}\stdv{0.5} & \bf{90.7}\stdv{1.7} &\bf{85.3}\stdv{0.1}& \bf{89.2}\stdv{0.3}& \bf{83.2}\stdv{1.5}& \bf{88.6}\stdv{1.0} &90.9\stdv{0.2}& \bf{88.0}\stdv{0.8}\\
\midrule
\midrule
Supervised&92.8\stdv{0.8}& 94.4\stdv{0.5} &92.7\stdv{0.4}&91.5\stdv{0.9}&88.2\stdv{1.0}&89.9\stdv{0.5}&92.3\stdv{0.8}&91.3\stdv{0.7}\\
\bottomrule
\end{tabular}}
\label{tab:ImageNet30}
\end{subtable}
\label{tb:novel_det}
\end{table}

\subsection{OOD detection}
Since our \textit{OpenPrompt} could detect OOD samples from unlabeled data, we evaluated the performance of our prompt-driven joint space learning in separating ID samples from OOD samples in unlabeled data. Following~\cite{saito2021openmatch}, we let the following datasets as OOD samples: SVHN~\cite{netzer2011reading}, LSUN~\cite{yu2015lsun}, CIFAR100 and ImageNet for CIFAR10 experiments (see Table~\ref{tab:CIFAR10}), and LSUN, Dogs~\cite{khosla2011novel}, CUB-200~\cite{wah2011caltech}, Caltech~\cite{griffin2007caltech}, DTD~\cite{cimpoi2014describing} and Flowers~\cite{nilsback2006visual} for ImageNet-30 (see Table~\ref{tab:ImageNet30}). The separation between the ID and OOD samples is evaluated by AUROC. We train a model utilizing all labeled examples of ID samples to show the gap from a supervised model. As can be seen in Table~\ref{tb:novel_det}, \textit{OpenPrompt} outperforms the various SOTA baselines by $1.1\%$ on SVHN and $0.4\%$ on average on CIFAR10. On ImageNet-30, \textit{OpenPrompt} improved $2.4\%$ on Flowers and $0.9\%$ on average compared with OpenMatch. These results demonstrate that our proposed \textit{OpenPrompt} is more sensitive to the representation gaps between ID and OOD samples, which leads to a more robust OSSL framework while they are exposed to unlabeled data containing outliers.

\subsection{Analysis}\label{Sec.4.2}
\begin{figure*}[t]
    \centering
    \includegraphics[width=\textwidth]{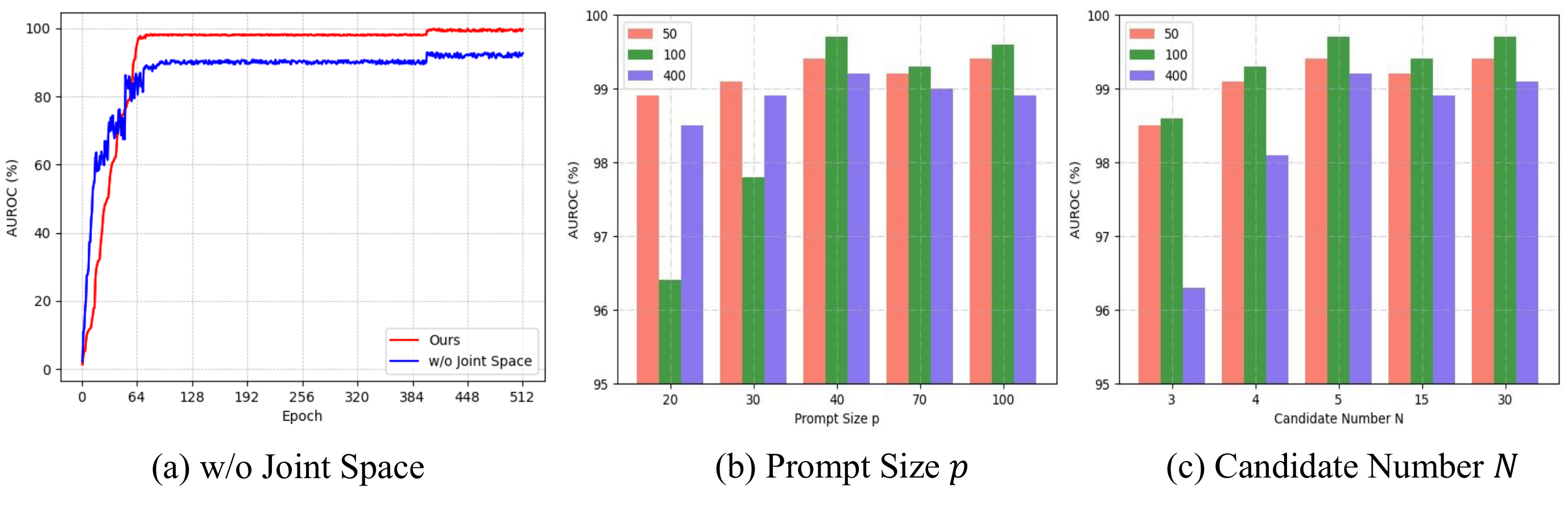} \vspace{-0.15cm}
    \caption{A set of ablative studies on CIFAR10. (a) The curve of the ablation study of joint space. (b) and (c) are the histograms of the different values of prompt size $p$ and candidate number $N$, where three colors represent the different numbers of labeled samples \textit{per class}.}
    \label{fig:ABS}
    \vspace{-0.6cm}
\end{figure*}

\xhdr{Effectiveness of prompt-driven joint space} To investigate the effectiveness of our proposed prompt-driven joint space learning mechanism, we built a model that uses a discriminator as an OOD detector to replace it (termed \textit{w/o} joint space). We record the AUROC values on CIFAR10 dataset with 100 labeled data per class and unlabeled data in Fig.~\ref{fig:ABS} (a), where the learning rate is decayed at epoch 400. As can be seen, our model converges faster and has higher accuracy (see the red curve). However, our method can still achieve satisfactory accuracy without the prompt-driven joint space mechanism (see the blue curve). This is likely because the visual prompt $v_{\phi}$ inherited from the labeled training stage is sensitive to the variability in the data space as it only has knowledge of the ID samples. However, the results are less than ours because the detector is randomly initialized, which will lead to significant oscillations in the early stages of training (see 10-70 epochs of the blue curve).

\xhdr{Effectiveness of prompt-wise contrastive representation} 
Here, we verify our claim that prompt-wise contrastive representation can further shape the joint space. As shown in the last two rows of Table~\ref{tab:benchmarks}, we use $~\textbf{\texttt{Ours}}$ (\textit{w/o} CL) to represent represents our model, but without prompt-wise contrastive learning (CL). As can be seen from this table, without the CL mechanism, the AUROC results of $~\textbf{\texttt{Ours}}$ (\textit{w/o} CL) will be lower than $~\textbf{\texttt{Ours}}$. When we derive the ID-specific and OOD-specific prompt for the prompt-wise contrastive representation, our method provides the highest AUROC values. It indicates that the prompt-wise contrastive representation can push the OOD samples away from the ID ones in unlabeled data, thereby enhancing the discrimination of outliers.

\begin{wraptable}{r}{0.65\textwidth}\footnotesize
\renewcommand{\arraystretch}{1.0}
\setlength{\tabcolsep}{2pt}
    \centering
    \caption{OOD detection results under different values of $\lambda$.} \vspace{-0.25cm}
    \begin{tabular}{lc| ccccccc}
\toprule
\multicolumn{1}{c}{$\lambda$} &&LSUN&DTD&CUB&Flowers&Caltech&Dogs&MEAN\\
\midrule
\multicolumn{1}{c}{0.1} &&88.1\stdv{0.2} &83.4\stdv{0.7}&86.5\stdv{0.8}&81.9\stdv{1.1}&87.3\stdv{0.5}&90.1\stdv{0.1}&86.2\stdv{0.5}\\
0.3 &&89.9\stdv{0.5}&84.7\stdv{1.5}&88.0\stdv{0.3}&{82.5}\stdv{2.1}&88.3\stdv{2.4}&90.6\stdv{1.7}&87.3\stdv{1.4}\\
\textbf{0.5} &&\bf{90.7}\stdv{1.7} &\bf{85.3}\stdv{0.1}& \bf{89.2}\stdv{0.3}& \bf{83.2}\stdv{1.5}& \bf{88.6}\stdv{1.0} &\bf{90.9}\stdv{0.2}& \bf{88.0}\stdv{0.8}\\
0.8 && {90.5}\stdv{0.9}& {84.9}\stdv{1.3}& {89.1}\stdv{1.8}& 82.7\stdv{2.9}& {87.9}\stdv{1.2}& {89.8}\stdv{3.4}& {87.5}\stdv{1.9}\\
1.0 && 90.4\stdv{2.2} &84.2\stdv{0.3}&88.9\stdv{0.8}&82.6\stdv{0.6}&88.5\stdv{1.1}&90.0\stdv{0.1}&87.4\stdv{0.9}\\
\bottomrule
\end{tabular}
    \label{tab:abls_od}
\end{wraptable}

\xhdr{Prompt discussion} 
We aim to find a balance between prompt size and model performance. Usually, it involves the template and size of the visual prompt. Following~\cite{bahng2022visual}, we choose the padding as the prompt template. Here, we analyze how the prompt sizes $p$ influence our method. As shown in Fig.~\ref{fig:ABS} (a), our model achieves the best AUROC scores at $p = 40$. When $p = 100$, the AUROC scores similar to the results of $p = 40$. However, the number of learnable parameters will be increased from 0.16M to 0.35M resulting in lower efficiency. 

\xhdr{Ablation study on $N$} As mentioned in \S\ref{Sec.3.3}, since the distribution of the unlabeled dataset is unknown in the pre-training stage, we should define multiple candidates for the initial binary classifier selection in the fine-tuning stage to find a better OOD center to construct the Apollonius circle. As can be seen in Fig.~\ref{fig:ABS} (c), the greater the number of candidates, the better results. However, the performance gradually stabilized when $N\geq 5$. It should be noted that we should compute the $\mathcal{A}_o$ values (see \S\ref{Sec.3.3}) for each candidate, thereby the greater the value of $N$, the greater the computational costs. Therefore, according to the result of Figure c, we set $N = 5$ in our experiments. 

\xhdr{Ablation study on $\lambda$} 
Here, we analyze the value of $\lambda$ how to influence the Apollonius circle. As shown in Table~\ref{tab:abls_od}, the larger value of $\lambda$, the greater AUROC performance of the OOD detection, \ie, $86.2$ ($\lambda = 0.1$) $\rightarrow$ $\textbf{88.0}$ ($\lambda = 0.5$). When $\lambda > 0.5$, the performances are no longer improving. Therefore, we set the $\lambda = 0.5$ to build the Apollonius circle.

\section{Conclusion}
In this paper, we propose a new efficient framework for OSSL (termed \textit{OpenPrompt}). Based on visual prompting, \textit{OpenPrompt} focuses on data space adaption instead of fine-tuning the whole model, which can achieve state-of-the-art performance with less than $1\%$ learnable parameters compared with other approaches. Our \textit{OpenPrompt} can effectively detect OOD samples thanks to the prompt-driven joint space learning mechanism that enlarges the distribution gap between ID and OOD samples. Furthermore, we reuse the OOD samples through prompt-wise contrastive representation to explore the structural information of ID and OOD samples and further shape the joint space. We believe this paper could inspire a shift in the direction of future research toward efficient OSSL.

\bibliography{iclr2023_conference}
\bibliographystyle{iclr2023_conference}
\clearpage
\appendix
\section{Appendix}
\subsection{Defination of Apollonius Circle}\label{ap:1}
\begin{figure*}[h]
    \centering
    \includegraphics[width=0.5\textwidth]{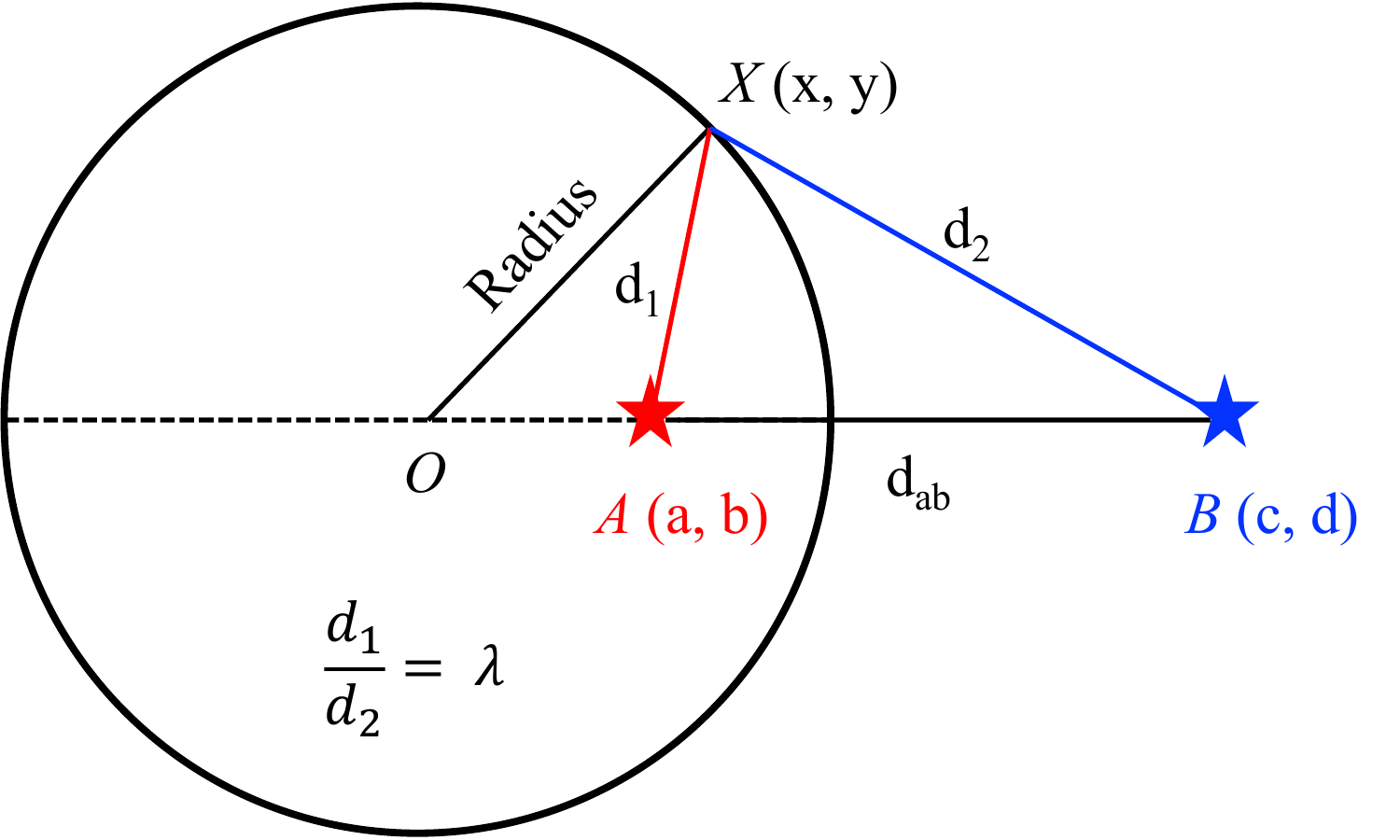} 
    \caption{Apollonius circle in a two-dimensional Euclidean space.}
    \label{fig:A_circle}
\end{figure*}
Given two points $A$ and $B$, Apollonius of Perga defines a circle as a set of points $X$ which satisfies the equation
\begin{equation}
    \mid XA \mid = \lambda \mid XB \mid,
\end{equation}
where $\lambda$ denotes a positive real number, $\mid XA \mid$ and $\mid XB \mid$ denotes the Euclidean distance from $X$ to $A$ and $B$, respectively. Assuming $X$, $A$ and $B$ are three points in a two-dimensional Euclidean space (see Fig~\ref{fig:A_circle}), we can define the radius of the Apollonius circle as follows:
\begin{equation}\label{eq:1}
    \begin{split}
    & \frac{\mid XA \mid}{\mid XB \mid} = \lambda \\
    & \Rightarrow \frac{d_1}{d_2} = \lambda \\
    & \Rightarrow \frac{\sqrt{(x-a)^{2} + (y-b)^{2}}}{\sqrt{(x-c)^{2} + (y-d)^{2}}} = \lambda \\
    & \Rightarrow (x - \frac{a-(\lambda)^{2}c}{1-\lambda^{2}})^2 + (y - \frac{b-(\lambda)^{2}d}{1-\lambda^{2}})^2 = \frac{(\lambda)^2}{(1-(\lambda)^2)^2} ((a-c)^2 + (b-d)^2)
\end{split}
\end{equation}
From the last row in Eq~\ref{eq:1}, we can define the radius of the Apollonius circle as follows:
\begin{equation}
    \begin{aligned}
      radius  & = \frac{\lambda}{1-(\lambda)^2} \sqrt{(a-c)^2 + (b-d)^2} \\
              & = \frac{\lambda}{1 - {\lambda}^2} d_{12}, \\
    \end{aligned}
\end{equation}
where $d_{12}$ denotes the Euclidean distance between point $A$ and point $B$ (as shown in Fig~\ref{fig:A_circle}).
\end{document}